\documentclass[conference]{IEEEtran}
\usepackage{times}
 \usepackage{booktabs}
\usepackage[numbers]{natbib}
\usepackage{multicol}
\usepackage{xcolor}
\definecolor{linkbrown}{RGB}{120,72,32}
\usepackage[
  bookmarks=true,
  colorlinks=true,
  linkcolor=linkbrown,
  citecolor=linkbrown,
  urlcolor=linkbrown
]{hyperref}
\usepackage{graphicx}
\usepackage{makecell} 
\usepackage{xspace}
\usepackage{caption} 
\newcommand{\algname}{\texttt{SoftAct}}

\usepackage{amsmath}
\usepackage{amssymb}
\usepackage{algorithm}
\usepackage{algpseudocode}
\usepackage{multirow} 
\usepackage{balance}
\usepackage{placeins}
\begin{document}

\title{Functional Force-Aware Retargeting from Virtual Human Demos to Soft Robot Policies}


\author{
\authorblockN{
Uksang Yoo\authorrefmark{1}\authorrefmark{2},
Mengjia Zhu\authorrefmark{2},
Evan Pezent\authorrefmark{2},
Jom Preechayasomboon\authorrefmark{2},
Jean Oh\authorrefmark{1}, 
Jeffrey Ichnowski\authorrefmark{1},\\
Amir Memar\authorrefmark{2},
Ben Abbatematteo\authorrefmark{2},
Homanga Bharadhwaj\authorrefmark{2},
Ashish Deshpande\authorrefmark{2},
Harsha Prahlad\authorrefmark{2}
}
\authorblockA{\authorrefmark{1}\textbf{The Robotics Institute, Carnegie Mellon University}}
\authorblockA{\authorrefmark{2}\textbf{Meta Reality Labs}} \\
\texttt{\href{https:\\soft-act.github.io}{soft-act.github.io}} \\\\
}

\makeatletter
\let\old@maketitle\@maketitle
\renewcommand{\@maketitle}{
  \old@maketitle
  \begin{center}
    \includegraphics[width=0.98\linewidth]{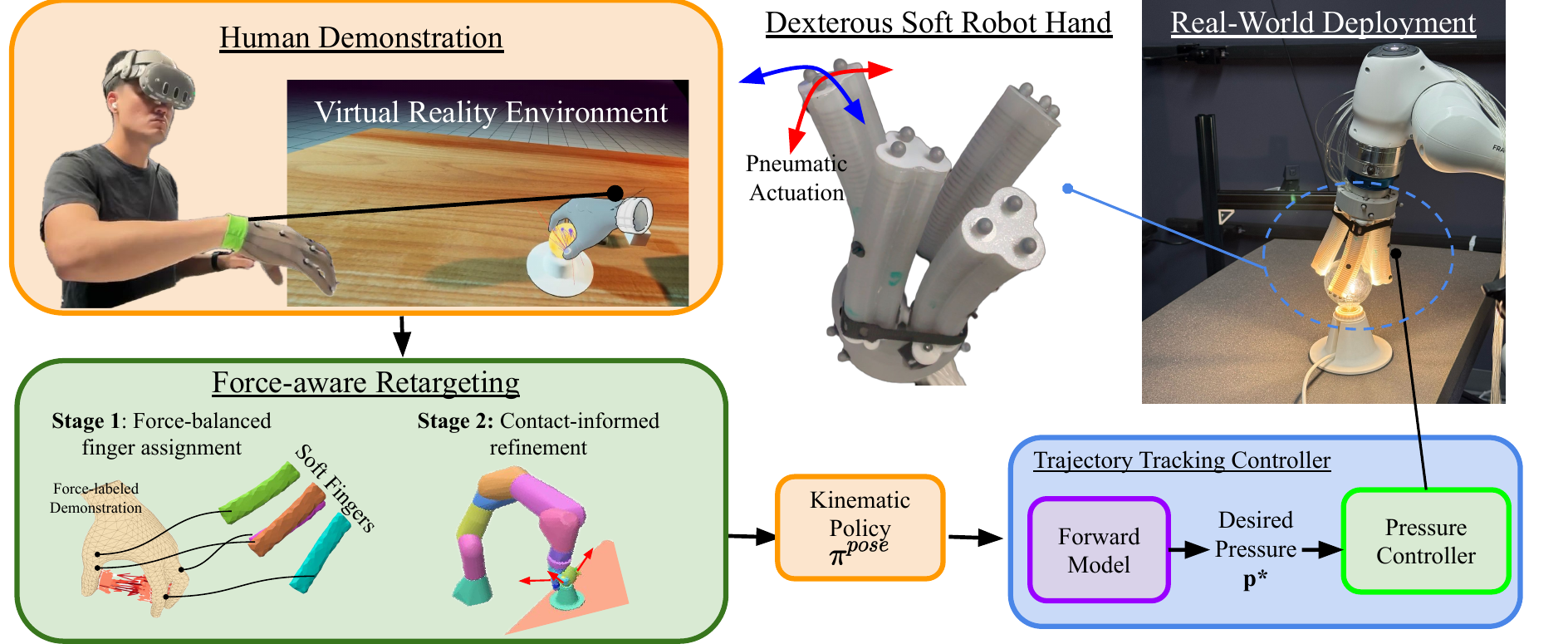}
    \captionof{figure}{
     \algname{} transfers human manipulation skills from VR demonstrations to a non-anthropomorphic pneumatic soft robot hand. The system uses contact force data to map human hand interactions onto the robot, enabling stable, functional manipulation across diverse tasks despite differences in hand structure and actuation. } 
    \label{fig:front}
  \end{center}
  \vspace{-0.12in}
}

\makeatother
\let\oldmaketitle\maketitle
\renewcommand{\maketitle}{
  \oldmaketitle
  \addtocounter{figure}{-1}
}

\maketitle

\begin{abstract}
We introduce \algname{}, a framework for teaching soft robot hands to perform human-like manipulation skills by explicitly reasoning about contact forces. Leveraging immersive virtual reality, our system captures rich human demonstrations, including hand kinematics, object motion, dense contact patches, and detailed contact force information. Unlike conventional approaches that retarget human joint trajectories, \algname{} employs a two-stage, force-aware retargeting algorithm. The first stage attributes demonstrated contact forces to individual human fingers and allocates robot fingers proportionally, establishing a force-balanced mapping between human and robot hands. The second stage performs online retargeting by combining baseline end-effector pose tracking with geodesic-weighted contact refinements, using contact geometry and force magnitude to adjust robot fingertip targets in real time. This formulation enables soft robotic hands to reproduce the functional intent of human demonstrations while naturally accommodating extreme embodiment mismatch and nonlinear compliance. We evaluate \algname{} on a suite of contact-rich manipulation tasks using a custom non-anthropomorphic pneumatic soft robot hand. \algname{}'s controller reduces fingertip trajectory tracking RMSE by up to 55\% and reduces tracking variance by up to 69\% compared to kinematic and learning-based baselines. At the policy level, \algname{} achieves consistently higher success in zero-shot real-world deployment and in simulation. These results demonstrate that explicitly modeling contact geometry and force distribution is essential for effective skill transfer to soft robotic hands, and cannot be recovered through kinematic imitation alone. Project videos and additional details are available at \href{https:\\soft-act.github.io}{soft-act.github.io}.

\end{abstract}

\IEEEpeerreviewmaketitle

\section{Introduction}

Soft robot hands offer intrinsic compliance, safety in human interaction, and robustness to uncertainty, making them particularly well suited for contact-rich manipulation tasks~\cite{yao2025soft, yoo2024moe, yoo2025soft, chen2022review, yoo2025kinesoft}. These advantages, however, come with fundamental challenges for policy learning and control~\cite{rus2015design}. In particular, translating dexterous manipulation skills from human demonstrations to soft robotic hands remains difficult due to extreme differences in workspaces, affordances, and contrability.

Unlike rigid and anthropomorphic robot hands, soft robotic hands are typically underactuated and driven with nonlinear actuators such as artificial muscles, exhibiting nonlinear, deformation-based motion rather than jointed kinematics~\cite{abondance2020dexterous, teeple2020multi}. Their morphologies and finger configurations often differ substantially from the human hand in degrees of freedom, workspace geometry, and contact affordances~\cite{alici2018softer}. As a result, direct correspondence between human and robot configurations is ill-defined, rendering classical imitation and retargeting approaches ineffective for many soft-hand systems.

 Prior work on learning from demonstration and hand retargeting often assumes a meaningful kinematic mapping and finger-to-finger correspondences between the demonstrator and the robot~\cite{sivakumar2022robotic,antotsiou2018task, pan2025spider, mandi2025dexmachina}. Human joint trajectories, fingertip positions, or full hand poses are mapped to robot configurations via inverse kinematics, heuristic rules, or learned pose-to-pose mappings~\cite{cheng2024open, qin2022one, qin2023anyteleop, li2025maniptrans}. While effective for anthropomorphic robot hands with similar structure, these methods break down for non-anthropomorphic and soft hands as we demonstrate in our experiments, where kinematic matching is often ill-posed and insufficient to reproduce functional manipulation, even when approximate geometric similarity is achieved.

Contact information can provide some guidance in generalizing manipulation strategies form human demonstrators to the robot embodiments~\cite{lakshmipathy2025kinematic, mandi2025dexmachina, li2025maniptrans}. However two limitations exist: obtaining such contact labeled datasets and these methods still require an explicit mapping between the hand surfaces to align the contact patches~\cite{lakshmipathy2025kinematic}, limiting their applications to mostly anthropomorphic target embodiments.

In this paper, we propose \algname, a \emph{contact force-guided} learning-from-demonstration framework for transferring human manipulation skills to non-anthropomorphic soft robotic hands. Our approach leverages immersive virtual reality to collect human demonstrations, capturing not only hand kinematics and object motion, but also dense contact patches and detailed contact force information throughout manipulation.
Rather than directly retargeting human joint trajectories, \algname{} introduces a two-stage force-aware retargeting algorithm. In the first stage, contact forces are attributed to individual human fingers, and robot fingers are allocated proportionally, establishing a force-balanced mapping between the human and robot hands. In the second stage, online retargeting is performed by combining baseline end-effector pose tracking with geodesic-weighted fine adjustments, using contact and force data to refine robot fingertip targets in real time. This formulation enables the robot to reproduce the functional demonstration by explicitly optimizing for geometry in contact points, while respecting the unique kinematic and morphological constraints of soft robotic hands.
By optimizing contact- and force-based objectives under the constraints of a pneumatically actuated soft hand, \algname{} naturally discovers robot-specific manipulation strategies that may differ substantially from human motion, yet achieve similar interaction outcomes. This stands in contrast to kinematics-driven retargeting methods, which enforce geometric similarity and often struggle under severe embodiment mismatch.

We evaluate our approach on multiple contact-rich manipulation tasks using a custom non-anthropomorphic pneumatic soft robot hand. Across tasks, our contact-centric retargeting produces more stable interactions and higher task success rates than kinematics-only baselines. Ablation studies further highlight the contributions of individual system components. The resulting behaviors exhibit distinct contact patterns and force distributions that reflect the unique affordances of pneumatic actuation, demonstrating that contact-level representations are essential for transferring human manipulation skills to soft robotic hands. Finally, we validate the simulated results with zero-shot real-world deployment, showing that \algname{} produces more stable and reliable rollout performances.

In summary, we make the following contributions:
\begin{itemize}
    \item A contact-centric learning-from-demonstration formulation for retargeting human manipulation skills to non-anthropomorphic soft robotic hands.
    \item A force-aware retargeting method that explicitly optimizes contact patches and force distributions rather than hardcoded kinematic similarity.
    \item A state-of-the-art learning-based low-level controller for tracking trajectories with the highly nonlinear pneumatic soft robot hand. 
    \item An end-to-end system for deploying the policy trained from virtual reality datasets on a pneumatically actuated soft hand, evaluated across multiple contact-rich tasks.
\end{itemize}

\section{Related Work}
We discuss prior works in dexterous manipulation with soft hands, learning from expert demonstrations, and human to robot motion retargeting. 

\subsection{Dexterous Soft Robot Manipulation}

Soft robotic hands have been widely studied for manipulation due to their intrinsic compliance, which enables passive adaptation to object geometry, robustness to uncertainty, and safe interaction with humans and fragile objects~\cite{rus2015design, chen2022review, zhou2023soft}. By distributing contact forces over extended surface areas, soft hands can form stable grasps and tolerate positioning errors that would lead to failure in rigid, jointed hands~\cite{abondance2020dexterous, teeple2020multi, liu2024skingrip}.

Prior work has improved the dexterity of soft hands through advances in actuator design, material composition, and bio-inspired morphologies~\cite{puhlmann2022rbo, firth2022anthropomorphic}. These developments have enabled increasingly sophisticated grasping behaviors and limited forms of in-hand manipulation, often relying on carefully hand-designed primitives and substantial expert tuning~\cite{puhlmann2022rbo}. Despite this progress, controlling soft hands remains fundamentally challenging due to their nonlinear, underactuated dynamics and deformation-driven motion, which complicate modeling, state estimation, and control~\cite{yasa2023overview}.

As a result, most learning-based approaches for soft robotic manipulation focus on grasp acquisition or stabilization rather than sustained dexterous manipulation~\cite{gupta2016learning}. Progress is further constrained by limited proprioceptive sensing and the difficulty of collecting high-quality demonstrations for soft hands~\cite{weinberg2024survey}. While the absence of rigid skeletal structures grants soft hands robustness and adaptability~\cite{pagoli2021soft}, it also introduces a severe mismatch between human hand kinematics and soft robot motion. Our work addresses this gap by moving beyond kinematic imitation and instead treating contact geometry and force as the primary transferable representations for dexterous manipulation.

\subsection{Learning from Expert Demonstrations for Manipulation}
Learning from demonstration (LfD) provides a powerful paradigm for transferring complex manipulation skills from humans to robots without explicit task modeling or reward specification~\cite{billard2019trends}. Demonstrations have been collected using a range of modalities, including kinesthetic teaching~\cite{zhang2025kinedex}, teleoperation~\cite{handa2020dexpilot, aldaco2024aloha, fu2024mobile}, motion capture, and video-based observation. These demonstrations are typically used to supervise imitation learning or behavior cloning policies, requiring aligned observation--action pairs from expert executions~\cite{chi2023diffusion, xu2025dexumi}.

LfD has been explored as a means to reduce the substantial manual effort required to program soft robot behaviors~\cite{yoo2025kinesoft,nazeer2023soft}. However, when there is significant embodiment mismatch between the human demonstrator and the robot, direct imitation of joint trajectories or hand poses becomes ill-posed. This challenge is especially pronounced for non-anthropomorphic and soft robotic hands, whose workspaces, degrees of freedom, and actuation mechanisms differ fundamentally from those of human hands. In such settings, geometric similarity alone is insufficient to capture the functional intent of the demonstrated manipulation.

\subsection{Human-to-Robot Motion Retargeting}

Motion retargeting aims to map human demonstrations to robot actions, often serving as an initial step for learning-from-human pipelines. Common retargeting methods rely on kinematic correspondence between human and robot hands, mapping joint angles, fingertip positions, or full hand poses using inverse kinematics, heuristic rules, or learned pose-to-pose mappings~\cite{qin2023anyteleop, lakshmipathy2025kinematic, yin2025geometric}. These approaches have shown strong performance for anthropomorphic robot hands with similar kinematic structures, but their effectiveness degrades significantly for non-anthropomorphic or soft hands.

To relax strict kinematic correspondence, recent work has explored task-space retargeting, object-centric alignment, and contact-aware formulations~\cite{pollard2004generalizing, lakshmipathy2025kinematic}. Some methods incorporate contact states or force-closure constraints to improve robustness, while others adopt optimization-based or policy-learning formulations to tolerate geometric mismatch~\cite{yin2025geometric, mandi2025dexmachina}. However, these prior works require explicit surface correspondence, and is primarily studied in the context of rigid, anthropomorphic hands~\cite{lakshmipathy2025kinematic, li2025maniptrans}.

Retargeting remains particularly challenging for soft and underactuated hands due to compliance, limited controllability, and nonlinear actuation~\cite{teeple2020multi, yasa2023overview}. Our work directly addresses these challenges by formulating retargeting as a contact- and force-centric optimization problem with simulation in the loop. By explicitly leveraging demonstrated contact patches and force distributions, our method enables soft robotic hands to reproduce the functional intent of human demonstrations while naturally exploiting their own morphology and actuation capabilities, rather than enforcing kinematic similarity.

\begin{figure}
\centering
\includegraphics[width=1.0\columnwidth]{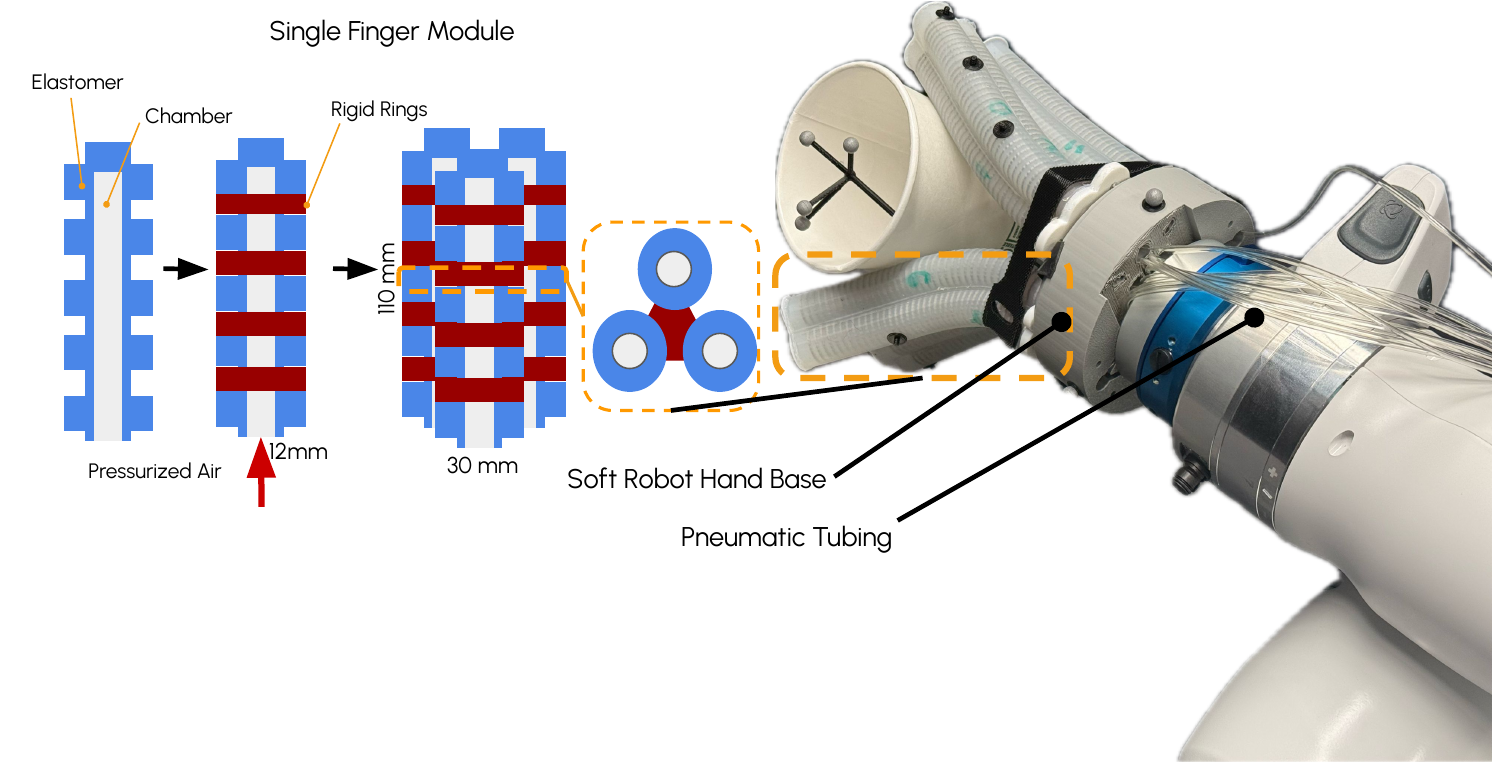}
\caption{\textbf{Design of the pneumatic soft manipulator.} The manipulator features a rigid–soft hybrid architecture, with each soft finger actuated by three radially arranged pneumatic chambers. The elastomeric finger body is reinforced with internal rigid structures to enhance controllability and repeatability. Differential pressurization enables planar and out-of-plane bending, while the fingers are mounted on a rigid base attached to a 7-DoF robotic arm.}
  \label{figure:design}
\vspace{-5mm}
\end{figure}

\section{SoftAct}
\label{sec:methods}

Mapping human manipulation demonstrations to soft robotic hands is challenging due to severe embodiment mismatch. Differences in morphology, actuation, and controllability cause naive kinematic retargeting to often fail to preserve the functional intent of manipulation.

We introduce \algname{}, a force-aware retargeting framework that transfers human demonstrations to soft-fingered robotic hands by explicitly reasoning over contact geometry and force distribution. Rather than enforcing pose-level correspondence, \algname{} treats contact forces as the primary transferable representation across embodiments.

Formally, given a human demonstration dataset
\[
\mathcal{D}_h = \{ (B_t, V_t, \mathcal{C}_t) \}_{t=1}^T,
\]
where $\mathcal{C}_t$ denotes hand--object contact locations and forces, our goal is to learn a robot control policy $\pi(a_t \mid o_t)$ that achieves functionally equivalent object interactions.

\begin{figure}[t]
\centering
\includegraphics[width=1.0\columnwidth]{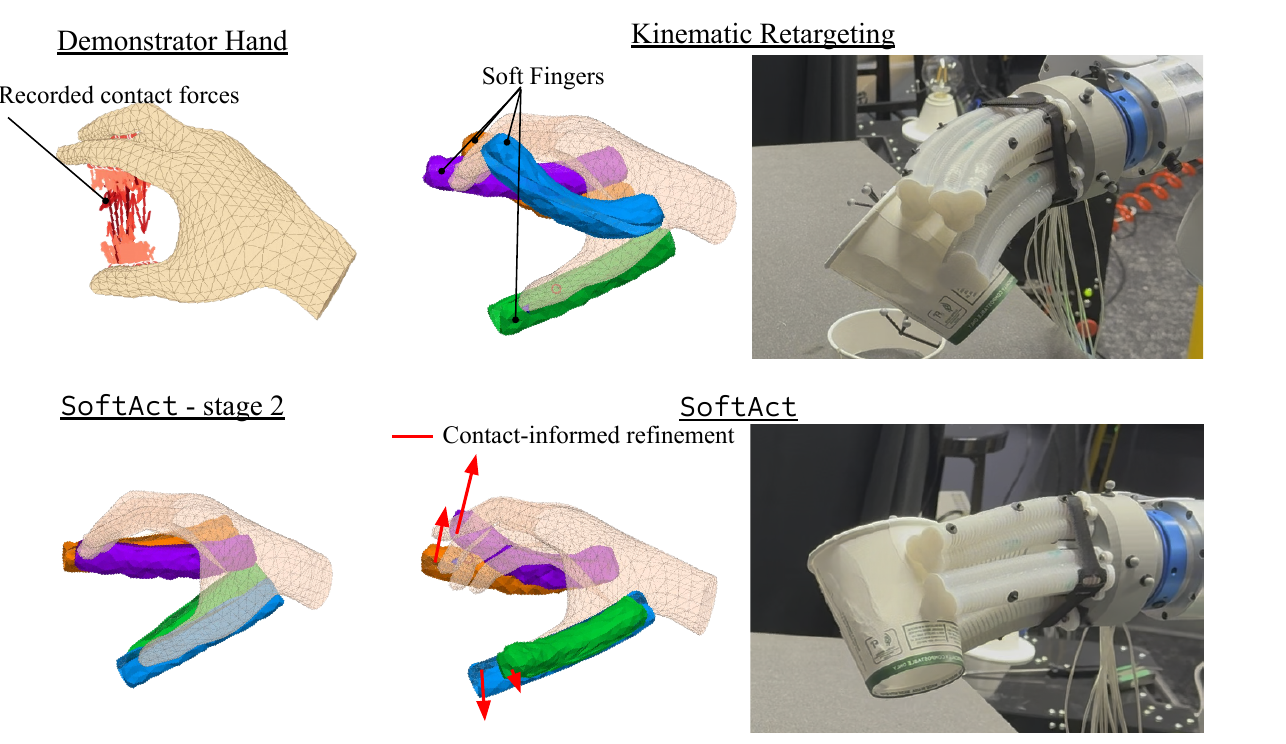}
  \caption{\textbf{Retargeting Stages.} Overview of the two-stage force-aware retargeting pipeline. Stage~1 performs offline force-balanced finger assignment, allocating robot fingers based on demonstrated contact force distribution. Stage~2 performs contact-informed refinement, adjusting fingertip targets using geodesic-weighted contact to align the contact surfaces between the human hand mesh and the soft robot.}
  \label{fig:stages}
\vspace{-5mm}
\end{figure}

\algname{} achieves this through a two-stage retargeting process as shown in Fig~\ref{fig:stages}. First, demonstrated contact forces are redistributed across robot fingers to produce a force-balanced human--robot mapping. Second, fingertip targets are refined online using geodesic-weighted contact influence on the hand surface, enabling stable, contact-consistent manipulation under extreme embodiment mismatch. The resulting policy is executed using a learned low-level controller for soft robotic actuation.

\subsection{Soft Robot Manipulator}

Each soft finger (Fig.~\ref{figure:design}) is actuated by three parallel pneumatic chambers arranged radially around the finger axis. Inspired by prior work on reproducible soft robot design and fabrication using 3D-printed internal structures~\cite{instant, yoo2021analytical}, we adopt a rigid--soft hybrid architecture that combines compliant elastomeric materials with embedded rigid elements to improve controllability and repeatability.

The finger body is cast from Ecoflex~00--30, providing high compliance and large deformation under modest pressure inputs. Internally, each finger contains a rigid strain-limiting structures that constrain undesired deformation modes such as ballooning and promote repeatable bending behavior. Differential pressurization of the three chambers produces planar bending and limited out-of-plane motion. The soft fingers are mounted on a rigid end-effector base attached to a 7-DoF robotic arm, allowing independent control of global end-effector motion and local finger deformation. The internal pressure to the soft robot chambers are controlled precisely at 200 Hz with a proportional pressure regulator.

\begin{figure}
\centering
\includegraphics[width=1.0\columnwidth]{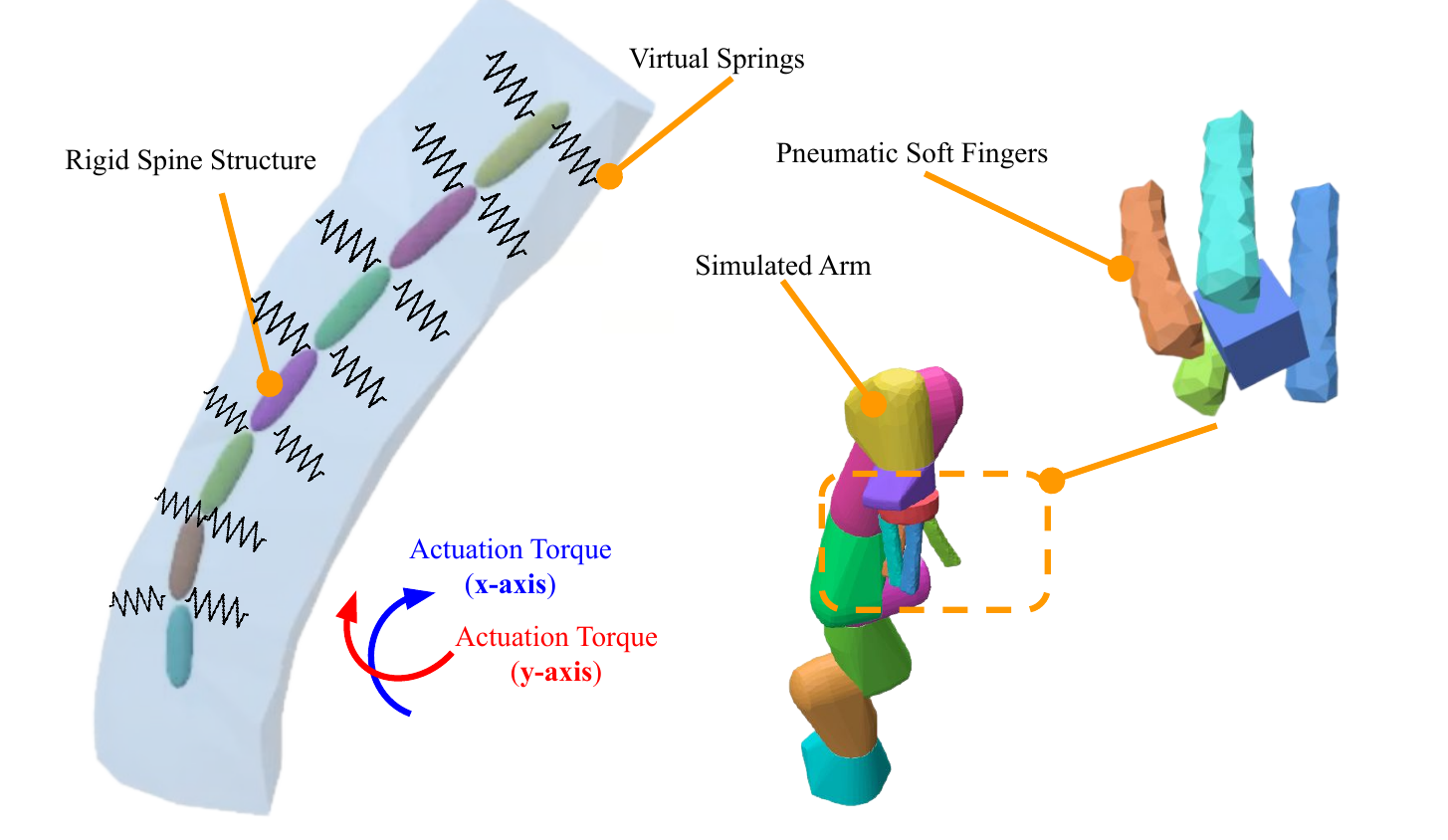}
  \caption{\textbf{Simulation Setup.} We approximate pneumatic pressure–induced bending using internal torque-based actuation applied to a rigid spine, which deforms the surrounding soft finger through distributed virtual spring constraints.}
  \label{fig:sim_setup}
\vspace{-5mm}
\end{figure}

\subsection{Simulation Environment}
\label{sec:methods-simulation}

Retargeting and evaluation are performed in a custom finite-element-based (FEM) physics engine that supports coupled rigid and soft body dynamics. As shown in Fig.~\ref{fig:sim_setup}, the simulated platform consists of a 7-DoF Franka Emika Panda arm equipped with a multi-fingered soft robotic hand, enabling contact-rich dexterous manipulation under controlled and repeatable conditions.

The Franka arm is modeled as a rigid-body manipulator, while each soft finger is represented as a deformable FEM body with material parameters chosen to approximate the mechanical behavior of the physical elastomer. To enable controllable deformation, pneumatic actuation is approximated in simulation through applied torques that induce articulated bending of the fingers. This abstraction captures the dominant deformation and contact behaviors relevant to manipulation while maintaining computational efficiency.

Together, this hybrid rigid–soft simulation framework provides a stable and expressive testbed for evaluating force-aware retargeting and contact-centric manipulation policies.

\begin{figure*}[t]
\centering
\includegraphics[width=\textwidth]{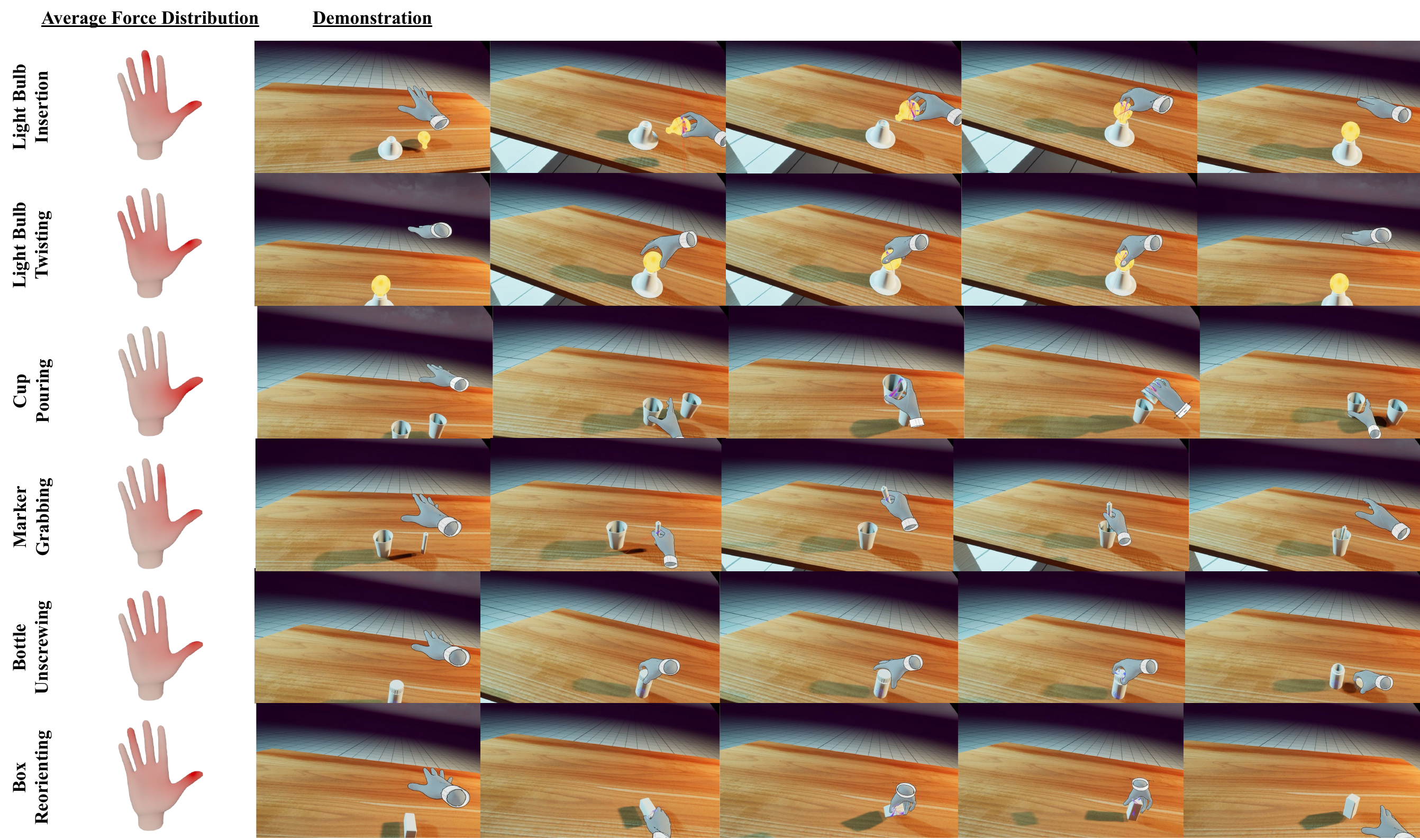}
\caption{
\textbf{Contact-Rich Manipulation Tasks and Demonstration Force Profiles.}
We evaluate \algname{} on six manipulation tasks (rows): \emph{light bulb insertion}, \emph{light bulb twisting}, \emph{cup pouring}, \emph{marker grabbing}, \emph{bottle unscrewing}, and \emph{box reorienting}. 
For each task, the left column visualizes the \emph{average demonstrated contact-force distribution} over the human hand, highlighting task-dependent asymmetries (e.g., thumb-dominant vs.\ distributed contact). 
The remaining columns show representative time-lapse frames from a single VR demonstration for the task.
}
\label{figure:tasks}
\vspace{-5mm}
\end{figure*}

\subsection{Human Demonstration Collection in Virtual Reality}

Human manipulation demonstrations are collected in an immersive virtual reality (VR) environment built using Unreal Engine. Unlike conventional VR-based teleoperation systems~\cite{qin2023anyteleop}, which use VR interfaces to directly control a physical robot in real time, our VR environment is used solely as a high-fidelity data collection platform for capturing rich human manipulation demonstrations.

The VR environment consists of a tabletop workspace that mirrors the geometry and scale of the real-world manipulation setup. Participants perform manipulation tasks using their right hand while interacting with virtual objects placed on the table.

Hand motion is captured using an OptiTrack motion capture system comprising ten cameras and 24 markers on the hand key points, providing high-fidelity tracking of the hand pose throughout each demonstration. The tracked hand pose is mapped in real time to a kinematic hand model in the VR environment, enabling accurate reproduction of hand motion and interaction geometry. Object poses are simultaneously recorded in a consistent world frame.

During each trial, we log full hand kinematics, object poses, and detailed interaction information at every timestep as shown in Fig.~\ref{figure:tasks}. Specifically, we extract contact patches between the hand and the object, represented as spatial distributions of contact points on the hand surface, along with the corresponding contact force magnitudes and directions. This rich contact-level information allows us to capture not only where contact occurs, but also how forces are applied and distributed over time.
By collecting demonstrations in VR, we obtain precise and noise-free measurements of contact geometry and force that are difficult to acquire reliably in the real world. These demonstrations serve as the primary input to our force-aware retargeting pipeline, enabling the transfer of human manipulation strategies to a non-anthropomorphic soft robotic hand.

\begin{figure}[t]
\centering
\includegraphics[width=1.0\columnwidth]{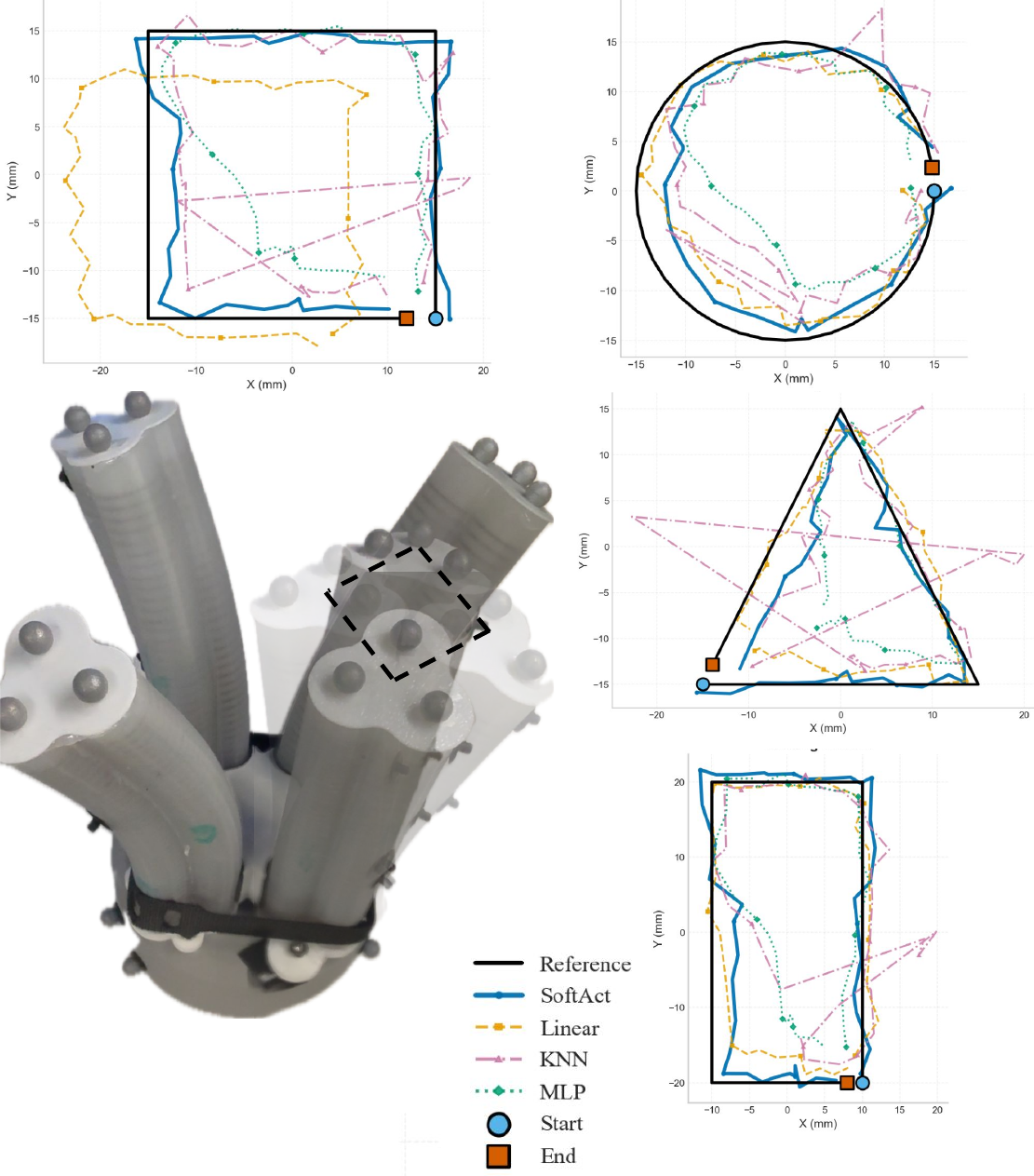}
\caption{\textbf{Low-level Control Performance.}
We evaluate trajectory tracking accuracy for a single soft finger executing planar reference trajectories, including square, circular, triangular, and rectangular motions. Each plot shows the desired fingertip trajectory in the $xy$ plane (black curve) and the executed fingertip trajectories produced by different controllers. Colored curves correspond to controller rollouts. Tracking error is computed as the Euclidean distance between the executed position and the reference trajectory at each timestep. The proposed controller produces trajectories that closely follow the reference paths with low bias and variance, whereas baseline controllers exhibit noticeable drift, distortion, and accumulated error.}
\label{fig:lowlevel}
\vspace{-5mm}
\end{figure}

\paragraph{Stage 1: Force-balanced finger assignment}

When transferring contact information across the hand, it is important that contact influence propagates along the physical surface of the hand rather than through free space. Using Euclidean distance in such cases would incorrectly couple physically unrelated regions of the hand.

To account for this, we measure distances intrinsically along the hand surface using mesh geodesics. Given two surface points $x$ and $y$, their geodesic distance is computed over the hand mesh graph,
\[
d_{\mathrm{geo}}(x,y)
=
\min_{P \in \mathcal{P}(\hat{v}(x), \hat{v}(y))}
\sum_{(i,j)\in P} \|v_i - v_j\|_2,
\]
ensuring that contact influence respects surface curvature and articulated geometry, which is critical for modeling interactions on soft fingers. Full implementation details are provided in the appendix.

Due to embodiment mismatch between the human hand and the soft robotic hand, direct one-to-one finger correspondence is generally infeasible. Instead, we estimate the cumulative contact load associated with each human finger over the demonstration and allocate robot fingers proportionally, allowing many-to-one mappings.

Contact forces are diffused over the hand surface using a geodesic heat kernel. For mesh vertex $v$ at timestep $t$, the heat value is defined as
\[
h_v^t
=
\sum_{(c,f)\in\mathcal{C}_t}
\|f\|_2 \exp\!\left(-\lambda\, d_{\mathrm{geo}}(v,c)\right).
\]
Heat values are accumulated over time and summed within each skeleton finger region to estimate the total load $F_r$ attributed to finger $r$.

Robot fingers are then allocated to balance contact load by solving
\[
\min_{\{n_r\}} \; \max_r \frac{F_r}{n_r}
\quad \text{s.t.} \quad \sum_r n_r = N_f,
\]
where $n_r$ denotes the number of robot fingers assigned to skeleton finger $r$, and $N_f$ is the total number of available robot fingers. The resulting assignment remains fixed throughout execution.

\paragraph{Stage 2: Retargeting with contact-informed refinement}
During execution, the robot follows a baseline end-effector trajectory while each soft finger $i \in \{1,\dots,N_f\}$ tracks its assigned human skeleton fingertip $\pi(i)$. To incorporate local contact interactions and align the contact surface of the human hand with the soft robot's contact surface, we apply a contact-informed adjustment computed from nearby demonstrated contacts.

At timestep $t$, demonstrated contacts are given by
$\mathcal{C}_t = \{(c_j^t, f_j^t)\}_{j=1}^{|\mathcal{C}_t|}$,
where $c_j^t \in \mathbb{R}^3$ denotes a contact location on the human hand surface and $f_j^t \in \mathbb{R}^3$ is the corresponding contact force.
Let $s_i^t \in \mathbb{R}^3$ denote the current fingertip position of robot finger $i$ expressed in the end-effector frame.

Each contact contributes a geodesic-weighted influence
\[
w_{ij}^t
=
\|f_j^t\|_2
\exp\!\left(
-\lambda\, d_{\mathrm{geo}}(s_i^t, c_j^t)
\right),
\]
where $d_{\mathrm{geo}}(\cdot,\cdot)$ denotes the geodesic distance on the hand surface.

The fingertip target adjustment is computed as a weighted average
\[
\delta_i^t
=
\frac{
\sum_j w_{ij}^t (c_j^t - s_i^t)
}{
\sum_j w_{ij}^t + \epsilon
},
\qquad
\|\delta_i^t\|_2 \le \delta_{\max}.
\]

This refinement enables each robot finger to adapt its target online based on demonstrated contact geometry and force distribution, while remaining anchored to the assigned human fingertip. Applying this procedure across the demonstration produces a set of functionally retargeted trajectories $\mathcal{D}_r$, which are subsequently used for imitation learning.


\begin{figure*}[t]
    \centering
    \includegraphics[width=\textwidth]{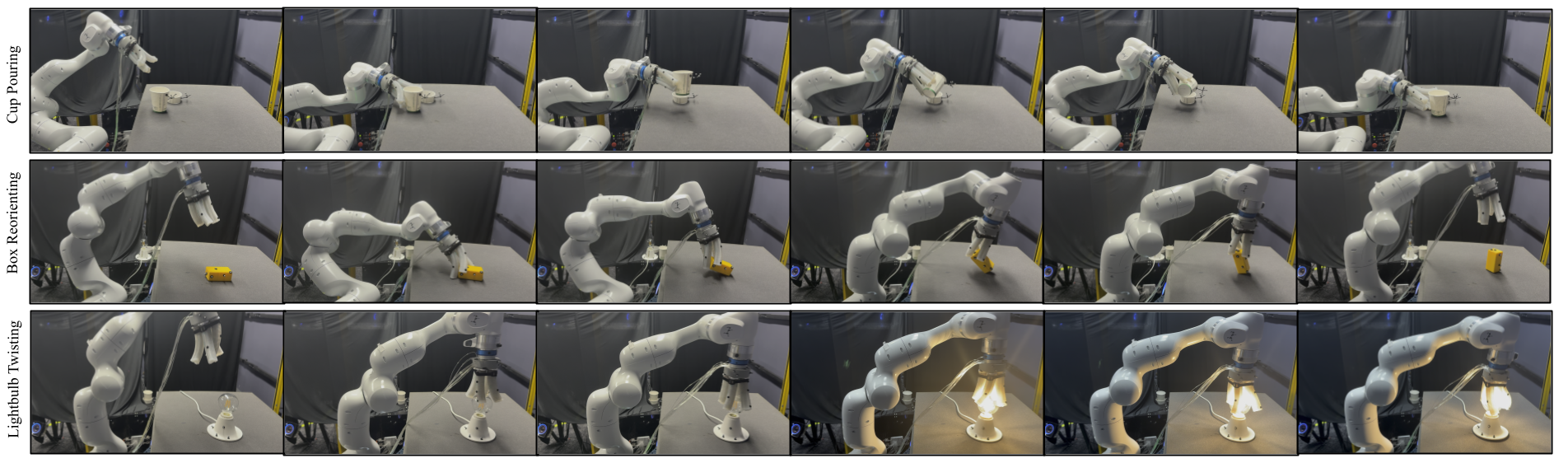}
    \caption{
    \textbf{Real-World Manipulation Results.}
    Sequential frames showing zero-shot real-world execution of contact-rich manipulation tasks with \algname{}, including paper cup pouring, box reorientation, and light bulb screwing. These tasks are challenging as they require compliant manipulation for contact-rich interaction. The robot successfully maintains stable contact and executes task-relevant object motion despite severe embodiment mismatch between the human demonstrator and the non-anthropomorphic soft robot hand. 
    }
    \label{fig:realworld}
    \vspace{-5mm}
\end{figure*}
\subsection{Policy Learning}
\label{sec:methods-policy}

We train a diffusion policy~\cite{chi2023diffusion} to imitate retargeted trajectories. The observation space is defined by robot proprioceptive state and object poses and the policy outputs a composite action consisting of an EE displacement, an EE orientation update, and soft-finger commands in torques as defined in Sec~\ref{sec:methods-simulation}.
Proprioceptive and object features are encoded with MLPs and concatenated before being passed to the diffusion model backbone. The policy outputs denoised actions at a fixed control rate, with all inputs and outputs normalized and clamped to system limits. During real-world deployment, the object poses are obtained with reflective motion capture markers placed on the surface of the objects; we leave vision-based sim2real for future work.

\subsection{Soft Finger Control}
\label{sec:methods-lowlevel}

Given the desired fingertip trajectories produced by the policy, we require a low-level controller capable of accurately tracking these targets using the available actuation degrees of freedom. Controlling soft robot fingers is fundamentally challenging due to their nonlinear, underactuated, and deformation-driven dynamics. Unlike rigid hands with well-defined joint coordinates, pneumatic soft fingers exhibit complex pressure–deformation relationships that are difficult to model analytically and vary across the workspace. As a result, direct kinematic inversion or linear control is insufficient for precise fingertip positioning. To address this, we introduce a learned low-level controller that maps actuator commands to fingertip motion.

Each soft finger is actuated with three pneumatic chambers. Inspired by prior works in learning-based nonlinear system control in other domains such as locomotion~\cite{sukhija2022gradient},  we learn a per-finger forward kinematic model using a MLP $f_{\mathrm{MLP}}: \mathbb{R}^3 \rightarrow \mathbb{R}^3,$ 
which maps chamber pressures $\mathbf{p} \in [p_{\min}, p_{\max}]^3$ to fingertip displacement $\mathbf{d} \in \mathbb{R}^3$ within safe pressure limits. This learned model captures the nonlinear compliant behavior of the soft fingers and is used by the inverse controller to compute pressure commands that track desired fingertip targets during execution.

At runtime, desired fingertip displacements $\mathbf{d}^{\mathrm{ref}}$, defined relative to a per-finger baseline configuration, are converted into pressure commands by inverting the learned forward model through optimization. For each waypoint, we solve
\begin{equation}
\mathbf{p}^{\mathrm{ref}} =
\arg\min_{\mathbf{p}\in [p_{\min}, p_{\max}]^3}
\left\|
\Pi_{xy}\!\left(f_{\mathrm{MLP}}(\mathbf{p})\right)
-
\Pi_{xy}\!\left(\mathbf{d}^{\mathrm{ref}}\right)
\right\|_2^2 ,
\label{eq:inv_opt}
\end{equation}
where $\Pi_{xy}(\cdot)$ extracts the planar $(x,y)$ components. Optimization is restricted to the planar subspace since all low-level tracking trajectories are planar, and out-of-plane motion is not independently controllable. 
The optimization problem in Eq.~\eqref{eq:inv_opt} is solved using gradient-based optimization through the differentiable MLP forward model.

To enforce actuator limits while preserving differentiability, pressures are re-parameterized using an unconstrained variable $\mathbf{u} \in \mathbb{R}^3$:
\begin{equation}
\mathbf{p}(\mathbf{u}) =
p_{\min} + (p_{\max} - p_{\min})\,\sigma(\mathbf{u}),
\label{eq:sigmoid_param}
\end{equation}
where $\sigma(\cdot)$ denotes the element-wise sigmoid function. This transforms the constrained optimization in Eq.~\eqref{eq:inv_opt} into an unconstrained problem over $\mathbf{u}$ while still allowing the optimizer to approach the full pressure range.

\paragraph{Input and output normalization.}
During training, both pressure inputs and displacement outputs are standardized. At inference time, the same normalization is applied inside the optimization loop:
\begin{equation}
\tilde{\mathbf{p}} =
(\mathbf{p} - \boldsymbol{\mu}_p) / \boldsymbol{\sigma}_p,
\qquad
\tilde{\mathbf{d}} =
(\mathbf{d} - \boldsymbol{\mu}_d) / \boldsymbol{\sigma}_d ,
\end{equation}
and the forward model is evaluated in normalized space before converting predictions back to metric units.

For trajectory execution, we generate a sequence of desired fingertip displacements
$\{\mathbf{d}^{\mathrm{ref}}_t\}_{t=1}^T$.
To improve convergence speed and stability, each optimization is warm-started from the previous solution. We solve the optimization using Adam for a fixed number of iterations and command the resulting pressures to the hardware at each waypoint.

\section{Experimental Evaluation}

We evaluate \algname{}'s force-aware retargeting and control approach  through a series of simulation and real-world experiments, assessing  low-level fingertip trajectory tracking, object-level trajectory tracking in simulation, and task success rates across a range of challenging, contact-rich manipulation tasks.

\subsection{Low-Level Trajectory Tracking}

The success of the policy rollout relies on accurate low-level controller performance to execute the predicted actions. To assess \algname{}'s controller performance, we first evaluate fingertip tracking accuracy on a single soft finger (Finger~0) executing planar square, circular, triangular, and rectangular trajectories. Each trajectory consists of 40 waypoints and is executed for three consecutive loops, with a dwell time of 0.5~s at each waypoint. Motion-capture feedback is used to obtain ground-truth fingertip positions for all trials.

We compare four controllers:
(i) \algname{}'s MLP-based inverse model optimized via gradient descent,
(ii) a linear Jacobian controller derived from local kinematic linearization inspired by ~\citet{sieler2023dexterous},
(iii) $k$-nearest-neighbor regression over sampled pressure–displacement pairs, and
(iv) direct MLP prediction of chamber pressures from desired fingertip displacements.

Table~\ref{tab:tracking_results_compact} reports trajectory tracking errors measured in the $xy$ plane and Fig.~\ref{fig:lowlevel} visualizes the resulting trajectories.
The learned \algname{} low-level soft robot pressure controller consistently achieves lower tracking error and reduced variance compared to all baselines.
Relative to the strongest baseline, \algname{} reduces RMSE by 55\% and reduces mean tracking error by 48\%.
Compared to learning-based baselines, the proposed controller reduces RMSE by 63\% and 74\%, respectively, and substantially decreases worst-case tracking error. These results indicate that optimizing a learned inverse model at execution time yields more accurate and stable trajectory tracking than direct prediction methods.
The improved precision and consistency support the use of \algname{} as a low-level execution module for contact-sensitive manipulation under nonlinear pneumatic actuation. 


\begin{table}[t]
    \centering
    \caption{Fingertip trajectory tracking performance for all controllers. Lower is better.}
    \label{tab:tracking_results_compact}
    \setlength{\tabcolsep}{4pt}
    \renewcommand{\arraystretch}{1.1}

    \begin{tabular}{lccc}
        \toprule
        \textbf{Controller} &
        \textbf{RMSE} &
        \textbf{Mean $\pm$ Std} &
        \textbf{Max} \\
        \textbf{} & \textbf{(mm)} & \textbf{(mm)} & \textbf{(mm)} \\
        \midrule

        Direct KNN      & 8.74          & 5.33 $\pm$ 6.93          & 34.34         \\
        Direct MLP      & 6.11          & 4.78 $\pm$ 3.80          & 13.85         \\
        Direct Linear   & 5.05          & 3.97 $\pm$ 3.13          & 10.89         \\
        \algname{}      & \textbf{2.28} & \textbf{2.06 $\pm$ 0.98} & \textbf{5.47} \\
        \bottomrule
    \end{tabular}

    \vspace*{-0.2cm}
\end{table}

\begin{table}[t]
\centering
\caption{Object trajectory tracking error in simulation experiments. Lower is better.}
\label{tab:sim_obj_tracking_ablation}
\setlength{\tabcolsep}{6pt}
\renewcommand{\arraystretch}{1.05}

\begin{tabular}{llcc}
\toprule
\textbf{Task} & \textbf{Method} & \textbf{Pos. (cm)} & \textbf{Rot. (deg)} \\
\midrule

\multirow{3}{*}{Light Bulb Insertion}
& Kinematic            & $1.52\pm0.61$          & $11.8\pm3.4$ \\
& \algname{}-stage 2   & $0.97\pm0.42$          & $25.1\pm7.9$ \\
& \algname{}           & $\mathbf{0.48\pm0.21}$ & $\mathbf{12.4\pm8.1}$ \\
\addlinespace

\multirow{3}{*}{Light Bulb Twisting}
& Kinematic            & $0.05\pm0.07$           & $18.2\pm6.5$ \\
& \algname{}-stage 2   & $0.07\pm0.03$           & $9.4\pm3.4$ \\
& \algname{}           & $\mathbf{0.02\pm0.005}$ & $\mathbf{2.9\pm1.3}$ \\
\addlinespace

\multirow{3}{*}{Cup Pouring}
& Kinematic            & $1.29\pm0.50$          & $12.5\pm4.1$ \\
& \algname{}-stage 2   & $0.81\pm0.35$          & $6.4\pm2.2$ \\
& \algname{}           & $\mathbf{0.51\pm0.21}$ & $\mathbf{3.5\pm1.3}$ \\
\addlinespace

\multirow{3}{*}{Marker Grasping}
& Kinematic            & $0.97\pm0.41$          & $14.3\pm5.0$ \\
& \algname{}-stage 2   & $0.56\pm0.25$          & $7.3\pm2.5$ \\
& \algname{}           & $\mathbf{0.31\pm0.13}$ & $\mathbf{3.9\pm1.5}$ \\
\addlinespace

\multirow{3}{*}{Bottle Unscrewing}
& Kinematic            & $0.26\pm0.11$          & $21.3\pm7.2$ \\
& \algname{}-stage 2   & $0.11\pm0.05$          & $11.1\pm3.8$ \\
& \algname{}           & $\mathbf{0.07\pm0.12}$ & $\mathbf{6.7\pm2.6}$ \\
\addlinespace

\multirow{3}{*}{Box Reorienting}
& Kinematic            & $1.34\pm0.54$          & $17.6\pm6.0$ \\
& \algname{}-stage 2   & $0.89\pm0.37$          & $9.0\pm3.1$ \\
& \algname{}           & $\mathbf{0.62\pm0.26}$ & $\mathbf{5.1\pm2.0}$ \\
\addlinespace

\bottomrule
\end{tabular}
\end{table}

\subsection{Object Trajectory Tracking in Simulation}

We next evaluate how improvements in low-level fingertip tracking translate to object-level manipulation performance in simulation. Executed object trajectories are compared against ground-truth demonstrations for paper cup pouring, light bulb insertion, and light bulb screwing. We compare three retargeting conditions: a kinematic-only baseline, \algname{}-stage~1 with force-balanced finger assignment only, and the full \algname{} method with contact-informed refinement. 

Table~\ref{tab:sim_obj_tracking_ablation} reports translational and rotational deviations across tasks. Across all evaluated settings, \algname{} consistently outperforms a kinematic-only baseline, reducing translational error by approximately 55--75\% and rotational error by up to 80\%. The largest gains occur in constrained insertion and rotational tasks, where maintaining stable contact is critical and kinematic retargeting accumulates significant drift.

While \algname{} with just stage 1 improves performance relative to pure kinematic retargeting, it remains consistently inferior to the full contact-aware formulation. This gap indicates that post-hoc adaptation alone cannot recover the benefits of explicitly reasoning about contact geometry and force distribution during retargeting.

\subsection{Policy Evaluation}

We evaluate policy-level performance on a suite of contact-rich manipulation tasks as listed in Fig.~\ref{figure:tasks} in both simulation and the real world. These tasks span asymmetric and distributed contact regimes, requiring stable force regulation, coordinated multi-finger interaction, and robustness to severe embodiment mismatch.

In simulation, \algname{} achieves consistent and substantial improvements over a kinematic-only baseline (Table~\ref{tab:success_rates}), increasing task success rates by roughly 30--70\% across manipulation settings. Gains are most pronounced for insertion and reorientation tasks, where kinematic policies frequently fail due to unstable or poorly distributed contact.

We further evaluate zero-shot real-world deployment, as shown in Fig.~\ref{fig:realworld}. 
\algname{} consistently outperforms the kinematic baseline across all evaluated tasks, achieving notable relative gains in real-world success rates. 
This suggests that explicitly reasoning about contact geometry and force distribution improves sim-to-real transfer for soft robotic manipulation.

Overall, the consistent gains in both simulation and real-world experiments indicate that explicitly modeling contact geometry and force distribution during retargeting is essential for reliable manipulation with non-anthropomorphic soft robotic hands, and cannot be recovered through kinematic imitation alone.

\begin{table}[t]
\label{tab:task_performance_sim_real}
\centering
\caption{Task success rates in simulation and the real world for \algname{} compared to a kinematic-only baseline. For each baseline per task, we do 30 rollouts in simulation, and 20 in the real world. Higher is better.}
\label{tab:success_rates}
\setlength{\tabcolsep}{6pt}
\renewcommand{\arraystretch}{1.05}
\begin{tabular}{llcc}
\toprule
\textbf{Task} & \textbf{Domain} & \textbf{\algname{}} & \textbf{Kinematic Baseline} \\
\midrule
Paper Cup Pouring     & Simulation & 90\%  & 40\%  \\
Light Bulb Insertion  & Simulation & 47\%  & 0\%   \\
Marker Grasping       & Simulation & 97\%  & 73\%  \\
Bottle Unscrewing     & Simulation & 80\%  & 47\%  \\
Box Reorienting       & Simulation & 90\%  & 30\%  \\
Light Bulb Screwing   & Simulation & 100\% & 77\%  \\
\addlinespace
Paper Cup Pouring     & Real       & 85\%  & 35\%  \\
Light Bulb Screwing   & Real       & 95\%  & 30\%  \\
Box Reorienting       & Real       & 70\%  & 10\%  \\
\bottomrule
\end{tabular}
\end{table}

\section{Conclusion}

In this paper, we presented a force-aware, contact-centric retargeting framework for transferring human manipulation skills to non-anthropomorphic soft robotic hands. Rather than enforcing kinematic correspondence between human and robot hands, our approach treats contact geometry and force as the primary transferable representations. By combining offline force-balanced finger assignment with online geodesic-weighted contact refinements, the method enables soft robotic hands to reproduce the functional intent of human demonstrations while naturally exploiting their own morphology and compliance.

We demonstrated the effectiveness of our approach across multiple contact-rich manipulation tasks, including paper cup pouring, light bulb insertion, light bulb twisting, bottle unscrewing, box reorienting and marker grasping. Quantitative results show that force-aware retargeting significantly outperforms kinematic-only baselines in both tracking accuracy and task success. Overall, this work suggests that contact- and force-centric representations provide a promising pathway for bridging extreme embodiment mismatch in imitation learning, enabling dexterous manipulation with soft robotic hands beyond the limits of kinematic correspondence.

\section{Limitations and Future Work}
Despite these results, several limitations remain. First, our current framework relies on access to ground-truth contact forces during demonstration, which can only be obtained in simulation. With recent progress in visual hand-object interaction understanding, extending the approach to operate with visually estimated contact forces would significantly broaden its applicability to real-world data collection. Additionally, our evaluation focuses on relatively short-horizon manipulation tasks with fixed interaction structure. Scaling the framework to longer-horizon, multi-stage manipulation remains an open challenge, particularly in settings where contact modes and functional roles evolve over time. Finally, for real-world deployment, we rely on object poses obtained from motion capture markers, which may not be available in many practical applications. An important direction for future work is to leverage the trained state-based policy as a teacher to train a vision-based student policy, enabling sim-to-real transfer without reliance on motion-capture markers.

\balance
\bibliographystyle{plainnat}
\bibliography{references}

\clearpage
\onecolumn
\appendices

\section{Simulation Environment and Soft-Hand Modeling}
\label{app:simulation}

This section describes the simulation environment used for retargeting and evaluation. The simulator is designed to capture the dominant deformation modes and contact interactions of the soft robotic fingers while maintaining numerical stability and computational efficiency. Rather than explicitly modeling pneumatic fluid dynamics, we adopt a hybrid rigid--soft abstraction that approximates pressure-induced bending through internal actuation torques.

\subsection{Soft Finger Model}

Each soft robotic finger is modeled as a deformable body using a stable Neo-Hookean material formulation~\cite{ding2022dynamic}. Material parameters are chosen to approximate the compliance of the physical elastomer while ensuring stable simulation under sustained contact-rich interaction.

\subsection{Rigid Spine Approximation}

To enable controllable deformation, each finger contains an internal rigid spine composed of a chain of capsule-shaped rigid bodies aligned along the finger’s longitudinal axis (Fig.~\ref{fig:sim_setup}). Adjacent capsules are connected via spherical joints, forming an articulated backbone that supports smooth bending under applied torques. Joint stiffness and damping are tuned to suppress high-frequency oscillations while preserving compliant motion.

\subsection{Soft--Rigid Coupling}

The rigid spine is coupled to the surrounding deformable mesh using distributed soft-to-rigid spring constraints. For each spine capsule, nearby FEM nodes within a fixed radius are connected via virtual springs. This distributed coupling acts as a tendon-like mechanism, transmitting spine motion to the soft body while avoiding localized stress concentrations and producing smooth, physically plausible deformation.

\subsection{Actuation Abstraction}

In the physical system, finger motion is generated through differential pressurization of three pneumatic chambers. In simulation, pneumatic actuation is abstracted by applying external torques to the rotational degrees of freedom of the spine capsules. Torques are applied about two orthogonal bending axes for all spine segments except the fixed base, inducing articulated bending that approximates the dominant deformation modes observed under pneumatic actuation. Explicit modeling of pressurized air dynamics is omitted to improve numerical stability and simulation throughput.

\subsection{Boundary Conditions}

Boundary conditions are applied at both the soft and rigid levels to anchor the hand. Soft-body nodes near the base of each finger are fixed in world coordinates, and the base capsule of each spine has all six degrees of freedom constrained. This prevents global drift and localizes deformation to the intended regions of the finger.

\subsection{Collision Handling}

Collision interactions are selectively enabled to balance physical realism and numerical stability. Self-collision within individual fingers and collisions between a spine and its enclosing soft body are disabled. Collisions between fingers and manipulated objects, as well as between spines of different fingers, are enabled to support realistic multi-finger contact during manipulation.

Multiple fingers are instantiated by arranging identical finger--spine assemblies around a circular base attached to the end-effector. Each finger is assigned a distinct collision group to allow fine-grained control over inter-finger interactions.

\subsection{Simulation Parameters}

Table~\ref{tab:sim_params} summarizes representative simulation parameters used across experiments. When exact physical measurements are unavailable, parameters are chosen to fall within commonly used ranges for soft-body manipulation.

\begin{table}[h]
\centering
\caption{Simulation parameters (representative values).}
\label{tab:sim_params}
\setlength{\tabcolsep}{6pt}
\renewcommand{\arraystretch}{1.05}
\begin{tabular}{lc}
\toprule
Parameter & Value \\
\midrule
Young’s modulus & 127 kPa \\
Poisson’s ratio & 0.48 \\
Material density & 1000 kg/m$^3$ \\
Spine capsule length & 10 mm \\
Number of spine segments & 8 \\
Joint stiffness & 0.2 N$\cdot$m/rad \\
Joint damping & 0.02 N$\cdot$m$\cdot$s/rad \\
Soft--rigid spring stiffness & 200 N/m \\
Simulation timestep & 1 ms \\
Contact friction coefficient & 0.6 \\
\bottomrule
\end{tabular}
\end{table}

\section{Surface Geometry and Distance Computation}
\label{app:geodesic}

This section describes the intrinsic surface distance computation used throughout \algname{} to propagate contact influence along the hand surface rather than through free space.

\subsection{Hand Surface Representation}

The hand surface is represented as a triangular mesh modeled as an undirected graph
$\mathcal{G} = (\mathcal{V}, \mathcal{E})$,
where vertices correspond to mesh vertices and edges connect adjacent triangles.

Given a 3D point $x$ on the hand surface, it is projected to the nearest mesh vertex
\[
\hat{v}(x) = \arg\min_{v_i \in \mathcal{V}} \|x - v_i\|_2 .
\]

\subsection{Geodesic Distance}

The geodesic distance between two surface points $x$ and $y$ is defined as
\[
d_{\mathrm{geo}}(x,y)
=
\min_{P \in \mathcal{P}(\hat{v}(x), \hat{v}(y))}
\sum_{(i,j)\in P} \|v_i - v_j\|_2,
\]
where $\mathcal{P}(\cdot,\cdot)$ denotes the set of all vertex paths on the mesh graph. Shortest paths are computed using Dijkstra’s algorithm with Euclidean edge weights.

This formulation ensures that contact influence respects surface curvature and articulated geometry, which is particularly important for highly curved and deformable structures such as soft fingers.

\section{Offline Force-Balanced Finger Assignment}
\label{app:offline}

This section corresponds to \textbf{Stage~1} of \algname{} (Sec.~III), which computes a fixed, force-balanced mapping between human and robot fingers prior to execution.

\subsection{Contact Representation}

At each timestep $t$, demonstrated contacts are represented as a set
\[
\mathcal{C}_t = \{(c,f)\},
\]
where $c$ denotes a contact location on the hand surface and $f$ is the corresponding contact force vector.

\subsection{Geodesic Force Diffusion}

Contact forces are diffused over the hand surface using a geodesic heat kernel. For mesh vertex $v$ at timestep $t$, the heat value is defined as
\[
h_v^t
=
\sum_{(c,f)\in\mathcal{C}_t}
\|f\|_2
\exp\!\left(-\lambda\, d_{\mathrm{geo}}(v,c)\right),
\]
where $\lambda$ controls the spatial decay of influence.

\subsection{Per-Finger Load Estimation}

Heat values are accumulated over time and summed over vertices associated with each skeleton finger region to obtain the total contact load
\[
F_r = \sum_t \sum_{v \in \mathcal{R}_r} h_v^t ,
\]
where $\mathcal{R}_r$ denotes the mesh region associated with skeleton finger $r$. When segmentation is unavailable, contacts are attributed to the nearest skeleton finger based on bone proximity.

\subsection{Load-Balanced Allocation}

Given $N_f$ available robot fingers, allocation is performed by solving
\[
\min_{\{n_r\}} \; \max_r \frac{F_r}{n_r}
\quad
\text{s.t.}
\quad
\sum_r n_r = N_f ,
\]
where $n_r$ denotes the number of robot fingers assigned to skeleton finger $r$.

\subsection{Workspace-Aware Finger Matching}

For each skeleton fingertip, we compute its demonstrated trajectory envelope in the end-effector (EE)--local frame. The envelope is aligned to the robot EE $xy$-plane via plane fitting followed by Procrustes alignment. Each robot finger workspace $\mathcal{W}_i$ is estimated through random pressure sampling.

The assignment cost between robot finger $i$ and skeleton fingertip $r$ is defined as
\[
C_{ir}
=
\|\mu_i - \mu_r\|_2
+
\beta\,\mathbb{I}\!\left[\mathcal{W}_i \cap \mathcal{W}_r = \varnothing \right].
\]
The optimal assignment is solved using the Hungarian algorithm, yielding a fixed mapping $\pi(i)$ that remains constant throughout execution.

\section{Online Contact-Informed Fingertip Refinement}
\label{app:online}

This section corresponds to \textbf{Stage~2} of \algname{} (Sec.~III), which performs online refinement of fingertip targets during execution.

\subsection{Geodesic Contact Weighting}

For each demonstrated contact $c_j^t$, we compute its geodesic distance to the current fingertip position $s_i^t$,
\[
d_{ij}^t = d_{\mathrm{geo}}(s_i^t, c_j^t),
\quad
w_{ij}^t = \|f_j^t\|_2 \exp\!\left(-\lambda\, d_{ij}^t\right).
\]

\subsection{Fingertip Adjustment Rule}

The adjustment vector is computed as
\[
\delta_i^t
=
\frac{\sum_j w_{ij}^t (c_j^t - s_i^t)}
{\sum_j w_{ij}^t + \epsilon},
\qquad
\|\delta_i^t\|_2 \le \delta_{\max}.
\]

The corrected fingertip target is tracked using Jacobian-based torque control in the end-effector frame.

\section{\algname{} Retargeting Algorithm}
\label{app:algorithm}

We summarize the full two-stage force-aware retargeting procedure used by \algname{}.

\FloatBarrier
\begin{algorithm}[H]
\caption{Force-Balanced Contact-Centric Retargeting}
\label{alg:force_balanced_retargeting}
\begin{algorithmic}[1]
\Require Demonstration frames $\{(B_t, V_t, \mathcal{C}_t)\}_{t=1}^T$
\Require Robot state $(p_{ee}^t, R_{ee}^t, \{s_i^t\}_{i=1}^{N_f})$
\Ensure Robot actions $\{a_t\}_{t=1}^T$

\State \textbf{Stage 1: Offline force-balanced finger assignment}
\State Compute geodesic distances on the hand surface
\State Accumulate contact force magnitudes per human finger
\State Allocate robot fingers to minimize maximum per-finger load
\State Solve workspace-aware assignment to obtain mapping $\pi(i)$

\State \textbf{Stage 2: Online retargeting with contact refinement}
\For{$t = 1$ to $T$}
    \State Compute baseline EE target $(\hat{p}_{ee}^t, \hat{R}_{ee}^t)$
    \For{each soft finger $i$}
        \State Compute geodesic-weighted contact influence
        \State Compute fingertip adjustment $\delta_i^t$
        \State Track skeleton fingertip $\pi(i)$ with adjusted target
    \EndFor
    \State Execute robot action $a_t$
\EndFor
\end{algorithmic}
\end{algorithm}
\FloatBarrier

\section{Training Details: Soft Finger Diffusion Policy}
\label{app:diffusion_training}

This section describes the training setup for the low-dimensional diffusion policy used to execute retargeted soft-finger trajectories.

\subsection{Policy Overview}

We train a conditional diffusion policy that predicts short-horizon soft-finger actions given recent robot observations. The policy operates in a low-dimensional action space and is conditioned on proprioceptive state and task-relevant features provided as a global conditioning signal.

\subsection{Sequence Structure}

Training is performed on fixed-length trajectory segments with a horizon of 16 steps. Each training sample includes the most recent 2 observation steps and predicts the next 8 action steps. No action latency is modeled, and past actions are not explicitly included in the observation stream.

\subsection{Model Architecture}

The policy backbone is a one-dimensional conditional UNet with three downsampling stages. The diffusion timestep embedding dimension is 256, with channel widths of \{256, 512, 1024\} and kernel size 5. Group normalization with 8 groups is used throughout. Conditioning is applied globally using concatenated observations across the temporal context window.

\subsection{Diffusion Process}

We employ a DDPM-style diffusion process with 100 diffusion steps during training and inference. The noise schedule follows a cosine-based schedule with variance clipping enabled. The model is trained to predict noise residuals (\(\epsilon\)-prediction).

\subsection{Optimization and Training Protocol}

Training uses the AdamW optimizer with learning rate \(1\times10^{-4}\), \(\beta=(0.95, 0.999)\), and weight decay \(1\times10^{-6}\). A cosine learning-rate schedule with 500 warmup steps is applied. Models are trained for 3000 epochs with batch size 256. An exponential moving average (EMA) of model parameters is maintained throughout training and used for evaluation.

\subsection{Inference}

At test time, action sequences are generated using 100 diffusion denoising steps and executed in a receding-horizon fashion.

\begin{table}[h]
\centering
\caption{Diffusion policy training hyperparameters.}
\label{tab:diffusion_hparams}
\setlength{\tabcolsep}{6pt}
\renewcommand{\arraystretch}{1.05}
\begin{tabular}{lc}
\toprule
Parameter & Value \\
\midrule
Horizon & 16 \\
Observation steps & 2 \\
Action steps & 8 \\
Diffusion steps & 100 \\
Batch size & 256 \\
Learning rate & $1\times10^{-4}$ \\
EMA decay & 0.999 \\
Training epochs & 3000 \\
\bottomrule
\end{tabular}
\end{table}

\clearpage
\section{Low-level Controller Training Data}
We visualize a downsampled dataset used for the low level soft robot finger controller used for real-world deployment.

\begin{figure}[h]
    \centering
    \includegraphics[width=0.57\linewidth]{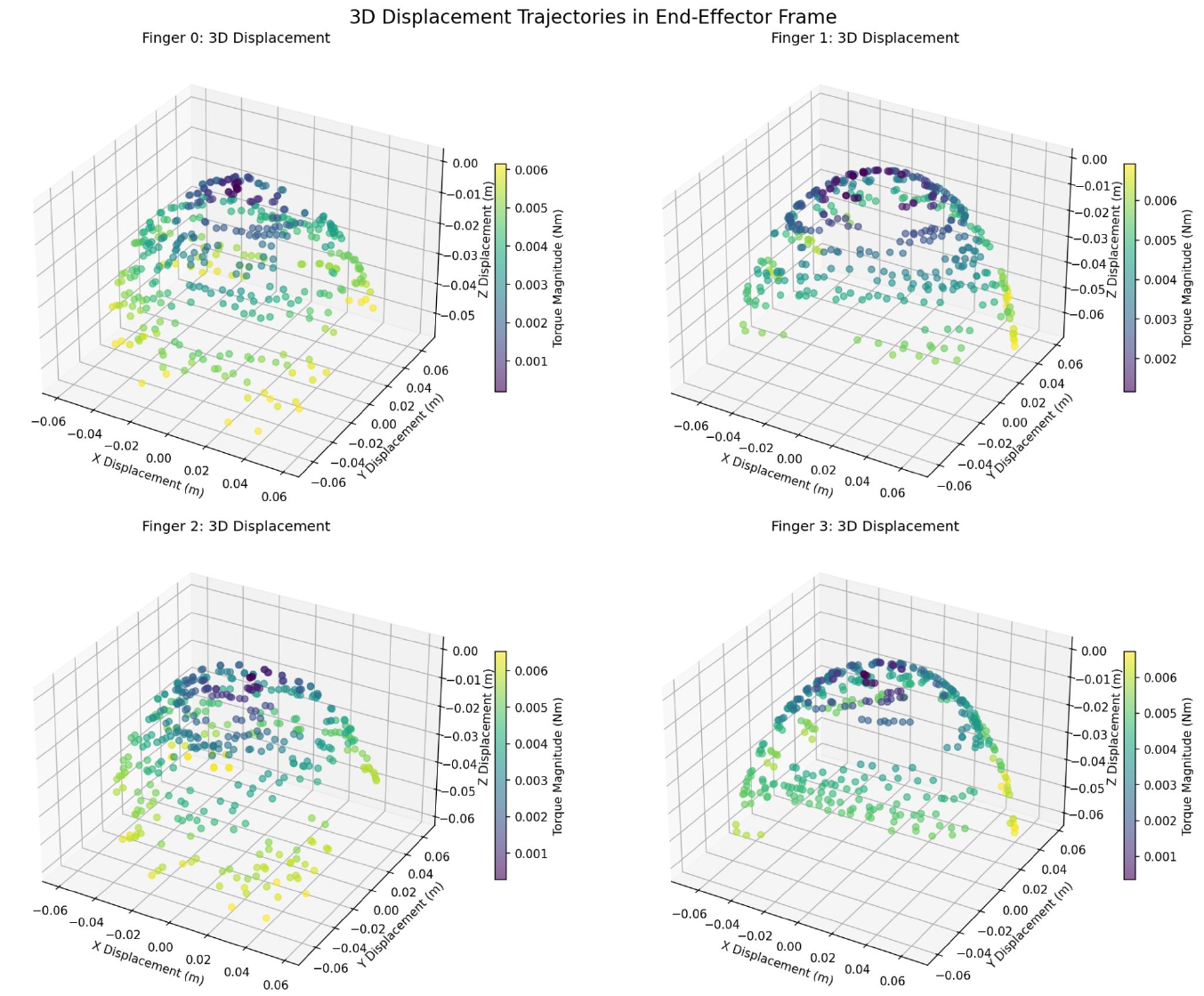}
    \caption{Down-sampled fingertip displacement data}
    \label{fig:enter-label}
\end{figure}

\end{document}